\documentclass[runningheads]{llncs}

\usepackage{accv}

\usepackage{accvabbrv}

\usepackage{graphicx}
\usepackage{booktabs}

\usepackage{colortbl}  %
\usepackage{pifont} %
\usepackage{enumitem} %
\usepackage[dvipsnames]{xcolor}
\usepackage{tikz}
\usepackage{subcaption}
\usepackage{xcolor}
\usetikzlibrary{positioning}

\usepackage{tcolorbox}

\usepackage[accsupp]{axessibility}  %

\usepackage{hyperref}

\usepackage{orcidlink}

\begin{document}

\title{Fine-Grained Generative Hallucination Detection and Correction for Aerial Image Captioning} 

\titlerunning{Fine-Grained Generative Hallucination Detection and Correction}

\author{Debolena Basak\inst{1}\orcidlink{0009-0001-6579-8397} \and
Somsubhra De\inst{2}\thanks{Work done during internship at IIT Hyderabad.}\orcidlink{0000-0003-3839-2575} \and
Maunendra Sankar Desarkar\inst{1}\orcidlink{0000-0003-1963-7338}
}
\authorrunning{Debolena B. et al.}

\institute{$^{1}$IIT Hyderabad, India \quad $^{2}$IIT Madras, India \\
\email{ai20resch11003@iith.ac.in, somsubhrade.04@gmail.com, maunendra@cse.iith.ac.in}}

\maketitle

\newcommand{\comr}[1]{\textcolor{red}{#1}}
\newcommand{\comb}[1]{\textcolor{blue}{#1}}
\newcommand{\comm}[1]{\textcolor{magenta}{#1}}

\begin{abstract}
Aerial imagery captured by unmanned aerial vehicles (UAVs) plays a vital role in applications such as surveillance, disaster response, and large-scale environmental monitoring. Automatic generation of textual descriptions for such imagery is essential for reducing manual inspection overhead and enabling better downstream decision-making. However, Multimodal Large Language Models (MLLMs), despite their impressive captioning capabilities, are prone to \textit{hallucination,} producing descriptions that are inconsistent with the visual input. This problem is significantly amplified in aerial imagery due to its unique challenges such as extreme viewpoint variation, small objects, class imbalance, and ambiguous visual entities. Existing hallucination detection approaches either rely on computationally expensive large external models such as GPT-4, which perform poorly on aerial domains, or treat the problem as a rigid binary classification task, ignoring the inherent uncertainty in ambiguous cases. To address these limitations, we propose a lightweight, reference-free framework for hallucination detection and correction in free-form aerial image captioning. Our method leverages state-of-the-art MLLMs, specifically InternVL3 and Qwen2.5-VL, to detect hallucinated content in generated captions with minimal computational overhead. Detected hallucinations are corrected through a dedicated refinement module that improves the factual consistency of the original captions. Extensive experiments on the Labelled Illusion Dataset (LID) demonstrate that our framework significantly reduces hallucination rates while remaining practical for real-time deployment in resource-constrained environments. Code is available at: \url{https://github.com/Debolena7/Hal_Detect}.

\end{abstract}

\section{Introduction}
\label{sec:intro}

\begin{figure}
\centering
\includegraphics[width=0.9\textwidth]{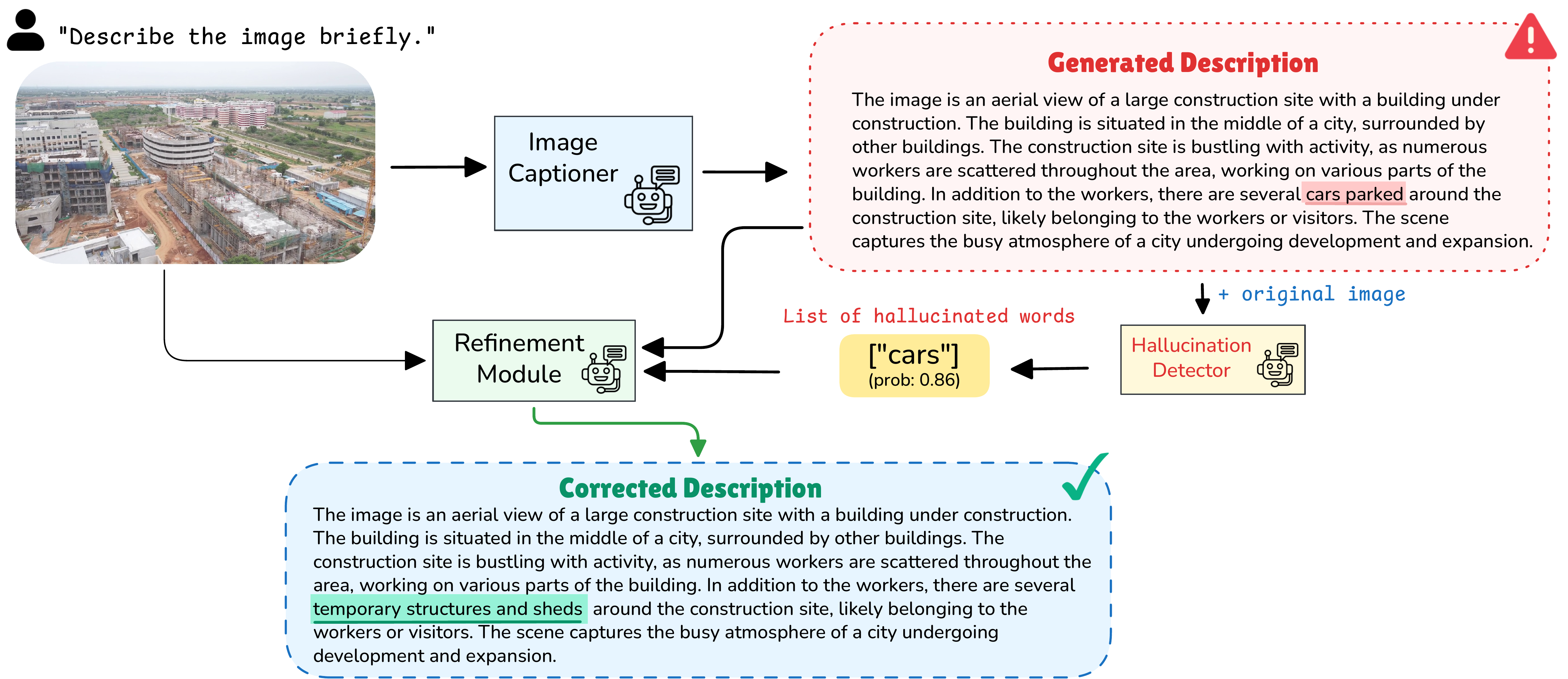}
  \caption{\small Overview of our proposed Hallucination Detection and Refinement Framework}
  \label{fig:framework}
  \vspace{-5mm}
\end{figure}

Drones are being increasingly adopted for large-scale visual data acquisition, enabling efficient capture of aerial imagery in environments that are difficult or unsafe for humans to access. They are used in a wide range of applications, including surveillance, crowd monitoring, disaster response, and land surveying. In such scenarios, the ability to automatically interpret visual data and generate textual descriptions is critical for reducing reliance on manual inspection of large image repositories. This capability facilitates downstream tasks such as anomaly detection, incident identification, path planning, and large-scale data indexing, enabling timely and informed decision-making. Despite significant advances in deep learning and transfer learning for visual understanding, accurately describing aerial imagery remains inherently challenging due to variations in viewpoint, object scale, orientation, and presence of small or ambiguous visual entities.

Recent advances in instruction-tuned Multimodal Large Language Models (MLLMs), including GPT-4~\cite{GPT4}, LLaVA~\cite{llava_2023_neurips,llava_2024_CVPR}, InstructBLIP~\cite{instructblip}, MiniGPT4~\cite{minigpt4}, Qwen2.5-VL~\cite{qwen2.5vl}, and InternVL3~\cite{internvl3} have introduced new paradigms for understanding and describing visual content in natural language. These models demonstrate strong capabilities in general-purpose image captioning and reasoning tasks. However, their practical deployment is hindered by the issue of \textit{hallucination}, where generated outputs include content not grounded in the input image. Such errors are particularly problematic in high-stakes applications, as they can propagate misinformation and undermine user trust. Consequently, hallucination detection and mitigation have emerged as a critical research challenge, with several prior works~\cite{POPE, NOPE, survey_LVLMS, gavie_robust_instruction_tuning, llava-rlhf} investigating methods to improve output faithfulness.

The challenges associated with hallucination detection are further exacerbated in aerial imagery~\cite{aerial_mirage}. 
Aerial-view datasets exhibit significant variability in object scale, orientation, and spatial distribution, contain noise, and often feature tiny objects. This hinders reliable object recognition and attribute grounding, thereby increasing hallucinations in image descriptions~\cite{aerial_mirage}. Additionally, such datasets~\cite{DOTA, visdrone, aerial_mirage} often exhibit pronounced imbalance in both object sizes and class distributions, which adds to the difficulties~\cite{aruba-sairam-wacv23}. For example, the class distribution in the Labelled Illusion Dataset (LID)~\cite{aerial_mirage} has a very long tail, as observed in \cref{fig:top15_hal}. These factors collectively make hallucination detection in aerial image captions a particularly challenging and underexplored problem, emphasizing the strong need for its accurate detection.

Existing approaches to hallucination detection frequently rely on large external models, such as GPT-4~\cite{GPT4}, to verify generated outputs~\cite{LURE_ICLR24, gavie_robust_instruction_tuning,llava-rlhf, faithscore}. Although GPT-4~\cite{GPT4} and GPT-5~\cite{gpt52} are effective for general-purpose datasets, they perform poorly on hallucination detection in aerial imagery (\cref{tab:main_results}) due to it's extreme and inherent challenges as discussed above. Moreover, these methods incur substantial computational overhead and latency, limiting their applicability in real-time or resource-constrained environments. Also, continuous stable internet connectivity is required for API calls of these models, which is impractical for drones in remote areas.
Consequently, reliably detecting these objects requires specialised models trained on aerial image datasets.
In addition, most prior work treats hallucination detection as a binary classification problem, labelling outputs as either hallucinated or non-hallucinated. This formulation overlooks the inherent uncertainty in many cases, where the distinction between faithful and unfaithful generation is ambiguous.

To address these limitations, there is a critical need for efficient hallucination detection methods from drones or UAVs, that operate at lower computational cost while providing probabilistic assessments of hallucination. Such approaches enable more informative and reliable interpretation of model outputs by capturing varying degrees of confidence. For example, this can allow systems to alert users by flagging content that requires further scrutiny without discarding it outright, and 
to allocate additional verification steps or resources to outputs that are more likely to be hallucinated.
These capabilities are particularly important in complex and safety-critical or strategic domains, where decisions must balance accuracy, uncertainty, and computational efficiency.

To this end, we propose a reference-free framework for hallucination detection and correction in free-form aerial image captioning. Our approach leverages pretrained popular MLLMs, including InternVL3~\cite{internvl3} and Qwen2.5-VL~\cite{qwen2.5vl}, to develop efficient hallucination detection models for aerial imagery. 
Our framework operates in a generator-agnostic manner, making it compatible with any existing MLLM that produces image descriptions, thereby enhancing their factual reliability without sacrificing efficiency. 
Given an input image and its corresponding generated description, the proposed framework identifies hallucinated content with low computational overhead. The detected hallucinations are subsequently refined through a correction module that revises the original responses to improve factual consistency. Despite the simplicity of the task design, our best-performing method achieves a significant reduction in hallucination rates 
with an \textbf{ALOHa improvement of 6.1\%, and reductions of 60.7\% in CHAIRi and 58.8\% in CHAIRs} (compared to the original uncorrected captions of LID), enhancing the reliability of MLLMs for real-time deployment without incurring substantial computational cost.
Unlike prior works~\cite{halloc_cvpr25, metatoken} that frame hallucination detection as a binary classification problem, we pose it as a generative prediction task, where the model directly predicts hallucinated words from a given image-caption pair. 
We demonstrate in \cref{sec:ablation} that our generative formulation of hallucination detection outperforms the classification approach using the same model backbones on the same LID dataset.
This choice is motivated by evidence \cite{genRM_ICLR25, aloha_naacl24} that generative formulations are better  than discriminative approaches, as they leverage LLMs' native strengths, such as reasoning and instruction tuning, to capture subtle semantic inconsistencies between image and caption that binary labels fail to express. In our setting, this enables the detector to holistically reason over both visual content and caption semantics to pinpoint specific hallucinated words or phrases, making it particularly well-suited for the nuanced and visually complex domain of aerial imagery. 
In summary, our contributions in this work are:
\begin{itemize}[leftmargin=*, topsep=0.4em]
    \item We pose hallucination detection as a generative prediction task, where given an image-caption pair, the model should predict fine-grained hallucinated words or spans from the caption. We also show that it performs better than the classification approach for hallucination detection under our experimental settings.

    \item We develop efficient and lightweight hallucination detection models for aerial imagery with low computational overhead. 
    
    \item We propose a reference-free framework for hallucination detection and subsequent caption editing using prompting-based correction. Leveraging the detected hallucinated spans, our approach refines the initial captions, leading to consistent improvements in CHAIR~\cite{chair_paper} and ALOHa~\cite{aloha_naacl24} scores over the original descriptions in Labelled Illusion Dataset (LID)~\cite{aerial_mirage}. 
\end{itemize}

\section{Related Works}

Early work on hallucination detection primarily focused on identifying non-existent objects in generated descriptions. For example, CHAIR~\cite{chair_paper} adopts a rule-based approach that compares mentioned objects with ground-truth annotations, making it dependent on reference data and limited to object-level errors. More recent methods leverage powerful off-the-shelf models for verification. GAVIE~\cite{gavie_robust_instruction_tuning} uses GPT-4~\cite{GPT4} to assign a single accuracy score to generated responses. FaithScore~\cite{faithscore} is a reference-free metric that  %
decomposes captions into atomic facts and verifies each against the input image to identify hallucinations. 
UNIHD~\cite{unihd} performs hallucination identification and classification using a pipeline of large models%
. However, these approaches operate at a coarse granularity level and do not localize hallucinations, which is the focus of our work.

Recent works like MetaToken~\cite{metatoken} use a binary classifier at the token level to predict whether each token is hallucinated or not, but it is restricted to object hallucinations only, whereas our method detects hallucinations beyond objects.
It is also not generator-agnostic, as it requires access to the caption generator's internal token probability distributions and related decoding statistics, and has not been evaluated in domain-specific settings such as aerial imagery.
Moreover, the lack of publicly available code limits reproducibility.
HalLocalizer~\cite{halloc_cvpr25}, a classification-based hallucination detection model, represents a strong step toward fine-grained hallucination localization, but it is trained and evaluated exclusively on general-purpose imagery and does not address the unique visual challenges of aerial scenes such as small object sizes and angular viewpoints.
GLSim~\cite{glsim} and SVAR~\cite{devils_svar}
provide valuable interpretability insights as hallucination signal, but they are intrinsically model-dependent, requiring access to internal hidden representations/attention maps of the caption-generating model. 

In contrast, our approach targets the challenging aerial domain and operates in a generator-agnostic manner: given only a caption-image pair, our fine-tuned models (InternVL3 and Qwen2.5-VL) generatively predict hallucinated words without accessing internal model states, followed by a correction module that resolves identified hallucinations. This makes our framework practically deployable in real-world UAV scenarios where the caption-generating model may be a black-box or computationally inaccessible system.

\section{Fine-Grained Multimodal Hallucination Detection and Correction}
\label{sec:method}

\textit{Hallucination} refers to any word or phrase in an image description, such as objects, counts, positions, attributes, relationships, or actions, that is not present in or supported by the visual content of the image.  This represents the most fundamental and critical form of errors affecting model reliability for real-time deployment. In this work, we propose fine-grained multimodal hallucination detection as a generative task. Using the detected hallucinated words, we correct the original descriptions using a refinement module. %
An illustration of our framework is given in \cref{fig:framework}. A formal description of the task is described below.

\subsection{Hallucination Detector}
\label{sec:hal_detector}

Formally, given an input image~$\mathcal{I}$, and its MLLM-generated description $\mathcal{D} =\{d_1, d_2, ....,d_n\}$, a multimodal small language model (MSLM) is finetuned using instruction prompt $\mathcal{S}$ to detect a set of hallucinated words $\mathcal{\widehat{Y}}=\{y_1, y_2, \dots, y_k\} \in \mathcal{D}$, conditioned on $\mathcal{I,D}$ and $\mathcal{S}$. 
Output $\mathcal{\widehat{Y}}$ is generated in  auto-regressive manner:
\begin{equation}
\label{eq:autoregressive_general}
p_\theta(\mathcal{\widehat{Y}} \mid \mathcal{I,S,D}) = \prod_{i=1}^{k} p_\theta\!\left(y_i \mid y_{<i},\, \mathcal{I,S},\, d_1, d_2, \dots d_n \right) \\
\end{equation}
where $(y_1, \dots, y_k)$ denotes hallucinated words.

Conditioned on the formulation in \cref{eq:autoregressive_general}, multimodal inputs ($\mathcal{I,S}, d_1, \dots d_n$) are first transformed into a unified sequence of embeddings within the MSLM for auto-regressive decoding. The detector then localizes hallucinated words or spans directly within $\mathcal{D}$, and the auto-regressive decoding naturally yields token-level probabilities $p_\theta\!\left(y_i \mid y_{<i},\, \mathcal{I,S},\, d_1, d_2, \dots d_n\right)$ for each predicted hallucinated word $(y_1, \dots, y_k)$, providing a confidence estimate alongside each localization, as a byproduct. 

\subsection{Refinement Module}
\label{sec:refinement}

Given the input image $\mathcal{I}$, original description $\mathcal{D} = \{d_1, d_2, \dots, d_n\}$, and the detected hallucinated words $\mathcal{\widehat{Y}} = \{y_1, y_2, \dots, y_k\}$, the refinement module produces a corrected description $\mathcal{\widehat{D}}$ by conditioning on all three inputs along with a system prompt $\mathcal{S'}$ containing explicit editing instructions:
\begin{equation}
\label{eq:refinement}
\mathcal{\widehat{D}} = \mathcal{M}\!\left(\mathcal{I,\, D,\, \widehat{Y},\, S'}\right)
\end{equation}
where $\mathcal{M}$ denotes a state-of-the-art advanced MLLM or MSLM. The model is instructed to 
revise only the hallucinated content identified in $\mathcal{\widehat{Y}}$ while preserving the 
factually consistent portions of $\mathcal{D}$, yielding a refined description $\mathcal{\widehat{D}}$ 
that is better grounded in the visual content of $\mathcal{I}$.

Although the primary refinement setting conditions on the image as the source of visual truth, we additionally experiment with a text-only correction setting, where the model corrects the description based solely on the original description and the detected hallucinated words, without access to the image:
\begin{equation}
\label{eq:refinement_text}
\mathcal{\widehat{D}}_{\text{text}} = \mathcal{M'}\!\left(\mathcal{D,\, \widehat{Y},\, S'}\right)
\end{equation}
where $\mathcal{M'}$ denotes an LLM, Small Language Model (SLM), or MLLM/MSLM with only text inputs. 
This setting serves as an \textit{ablation} to quantify the contribution of visual context to the refinement quality, since providing the image enables the model to replace 
hallucinated content with visually grounded alternatives, producing more complete and faithful descriptions. Isolating 
this effect allows us to assess how much of the improvement is attributable 
to visual grounding versus the predicted hallucinated words alone.

It is worth noting that our core contribution is framing fine-grained hallucination detection as a generative prediction task for aerial imagery in a generator-agnostic manner, which to our knowledge has not been explored before. The refinement module is a downstream component demonstrating the practical utility of the detected hallucinations, completing the end-to-end pipeline. Its simplicity by design: a lightweight, reference-free step requiring no additional training, makes it deployable with any capable MLLM. 
We experimented with multiple prompt formulations on training examples and selected the best.

\section{Experimental Setup}
\subsection{Models}
\label{sec:models}

\textbf{Hallucination Detector Models:}
For the hallucination detection task, we fine-tune popular lightweight instruction-tuned Multimodal Small Language Models (MSLMs) that offer strong vision-language understanding while maintaining fast inference speed, making them suitable for real-time UAV deployment.
Specifically, we finetune InternVL3-Instruct (1B, 2B, 8B)~\cite{internvl3} and Qwen2.5-VL-Instruct-3B~\cite{qwen2.5vl} using LoRA~\cite{hu2021loralowrankadaptationlarge}.
We select these model families for their strong multimodal capabilities and distinct architectural and training strategies.\\ 
\noindent\textbf{Refinement Module:} As described in \cref{sec:refinement}, we employ advanced pre-trained MLLMs or MSLMs to correct hallucinated content (if detected by our detector module) and produce refined descriptions via an instruction system prompt. We select model families with strong instruction-following and multimodal capabilities: 
Qwen3.5-(4B, 9B, 27B)~\cite{qwen35blog} and Gemma-4-it (E4B, 26B-A4B, 31B)~\cite{gemma_4_2026}. These are natively multimodal and instruction-tuned variants with built-in reasoning capabilities and configurable reasoning modes.
For the text-only refinement setting, we also experimented with GPT-5.2~\cite{gpt52} and Qwen3-(8B, 14B)~\cite{qwen3}. 
Since Qwen3.5 and Gemma-4 variants demonstrate a superior cost-performance 
trade-off, we extend their evaluation to both text-only and multimodal settings 
for a comprehensive comparison across input modalities. Together, these two 
architecturally distinct model families allow us to benchmark the effectiveness 
of our zero-shot refinement approach across diverse multimodal backbones. 

All refiners use the predictions of our best performing detector -- InternVL3-8B.
Additional details regarding model choices, implementation and instruction prompts are provided in the supplementary 
(\cref{supple_sec:model_choice_details}, \cref{supple_sec:implementation_details}, \cref{supple_sec:prompts_hal_detect}, \cref{supple_sec:prompts_refinement}, respectively.)

\subsection{Dataset}
\label{sec:dataset}
For our experiments, we need an aerial imagery hallucination benchmark comprising of image-description pairs manually annotated with fine-grained hallucinated words or spans. The only such publicly available dataset is the Labelled Illusion Dataset (LID)\footnote{\href{https://github.com/Debolena7/Aerial_Mirage}{Labelled Illusion Dataset}}~\cite{aerial_mirage}. Here, image-description pairs are manually annotated with fine-grained hallucinated words or spans, where each sample may contain 
one or more hallucinations. It contains 4884 train and 1232 test samples.

\subsection{Baselines}
\label{sec:baselines}

\subsubsection{For Hallucination Detection:}
Our evaluation is structured across five tiers, each serving a distinct purpose:

\noindent\textbf{Naive Baseline:} We include a degenerate baseline that predicts no hallucination for every input (i.e., all-None predictions). Given the significant class imbalance inherent in hallucination detection — where non-hallucinated words vastly outnumber hallucinated ones — this baseline establishes a lower bound and highlights the difficulty of the task beyond trivial class prediction.

\noindent\textbf{Zero-Shot MSLMs:} We evaluate InternVL3-Instruct (1B, 2B, 8B) \cite{internvl3} and Qwen2.5-VL-Instruct-3B~\cite{qwen2.5vl} in a zero-shot setting, selecting model sizes comparable to our fine-tuned detectors to ensure a fair comparison under similar inference budgets. These baselines directly measure the inherent hallucination detection capability of off-the-shelf models without any task-specific adaptation, thereby justifying \textit{the need for fine-tuning on aerial imagery.}

\noindent\textbf{Proprietary Models:} We evaluate GPT-5.4-nano and mini~\cite{gpt5.4_mini_nano_2026} in both standard instruction-following mode and with ``low'' reasoning enabled. While these models are not viable for real-time UAV deployment due to API dependencies and latency constraints, \textit{they serve as strong baselines.} We additionally include the ``low'' reasoning setting to assess whether even a small degree of chain-of-thought reasoning yields meaningful gains in fine-grained hallucination detection, without the full computational cost of unconstrained reasoning.

\noindent\textbf{Existing Method:} We compare against HalLocalizer~\cite{halloc_cvpr25}, a token-level classification approach for hallucination detection, which we fine-tune on LID's training set for a fair comparison. This directly contrasts our generative prediction paradigm against a discriminative detection baseline, evaluating whether framing detection as sequence generation offers advantages over binary token-level classification for aerial imagery. Details regarding HalLocalizer's adaptation and implementation are provided in the supplementary (\cref{supple_sec:implementation_details}).

\noindent\textbf{Hallucination-only samples:} 
To decouple localization performance from the easier ``None'' decision, we also evaluate on the subset of test samples that contain actual hallucinations. Further details are provided in the supplementary (\cref{supple_sec:results_hal_only}).

\subsubsection{Refinement Module Baselines:}
We establish two reference points to evaluate the quality of our refined descriptions. First, we compare our module's corrected descriptions against the original uncorrected ones in the LID test set, measuring the direct improvement from the correction step. 
Second, we construct an oracle baseline by prompting GPT-5.2~\cite{gpt52}, a strong proprietary model with excellent instruction-following, to correct the original descriptions using the \textit{ground-truth} hallucinated words, under both (1)~text-only and (2)~multimodal (image + text) settings. The text-only variant reflects ideal correction from linguistic context alone, while the multimodal variant additionally leverages visual grounding to replace hallucinated content with visually accurate alternatives. Since both operate on ground-truth rather than predicted labels, they serve as strong reference points for our refinement module, to assess how closely our end-to-end pipeline approaches ideal correction quality under each input modality.

\subsection{Evaluation Metrics}
\label{sec:eval}
All comparisons are performed on the test set. For the hallucination detector, 
we report Precision, Recall, and F1 scores with respect to the groundtruth hallucinated words, along with per-sample inference time (in seconds) to assess deployment efficiency.
The score computation considers full-span matching, normalization through lower-casing followed by lemmatization, and multiple labels per instance. Scores are averaged per sample and then averaged across the test set.
For the refinement module, we evaluate the quality of corrected descriptions 
using hallucination-specific metrics: ALOHa~\cite{aloha_naacl24} and 
$\text{CHAIR}_i$, $\text{CHAIR}_s$~\cite{chair_paper} as primary metrics, 
which directly measures object hallucination in generated captions. 
As no single metric in literature holistically covers all categories, we use complementary evaluations.
We report image-caption similarity metrics: CLIPScore~\cite{clipscore}, RefCLIPScore~\cite{clipscore}, and BERTScore~\cite{BertScore} (DeBERTa-xlarge-mnli, with and without IDF weighting). The metrics measure the semantic similarity of the refined captions to the input image (CLIPScore), to both the image and ground-truth captions (RefCLIPScore), and to the ground-truth captions alone (BERTScore). %
Further, we perform human evaluation on a subset of examples to judge the quality of our refined captions.
We also report average per-sample inference time for the refinement module.

\section{Results and Analysis}

\begin{table}[t]
\centering
\resizebox{0.5\textwidth}{!}{%
\small
\begin{tabular}{lcccc}
\toprule
\rowcolor{blue!10}
\textbf{Model} & \textbf{Precision$\uparrow$} & \textbf{Recall$\uparrow$} & \textbf{F1$\uparrow$} & \textbf{Time(s)$\downarrow$} \\
\midrule

All-None Baseline & 49.11 & 38.27 & 43.01 & - \\

\midrule
\multicolumn{5}{c}{\textit{Zero-shot Baselines}} \\
\midrule

Internvl3-1B & 46.78 & 37.25 & 41.48 & 0.0753 \\
Internvl3-2B & 49.11 & 38.27 & 43.01 & 0.0796 \\
Internvl3-8B & 55.61 & 45.16 & 49.84 & 0.2934 \\
Qwen2.5VL-3B & 33.55 & 48.83 & 39.77 & 0.5451 \\
GPT-5.4-nano & 34.43 & 62.49 & 44.39 & 4.3229 \\
GPT-5.4-mini & 39.63 & 71.92 & 51.10 & 4.2535 \\
GPT-5.4-nano (reasoning) & 60.00 & 68.69 & 64.05 & 5.2977 \\
GPT-5.4-mini (reasoning) & 64.78 & 76.22 & 70.04 & 5.1933 \\

\midrule
\multicolumn{5}{c}{\textit{Finetuned Models}} \\
\midrule

HalLocalizer~\cite{halloc_cvpr25} & 57.85 & 46.75 & 51.71 & 0.0980 \\
Qwen2.5VL-3B (ours) & 78.66 & 72.04 & 75.21 & 0.5569 \\
Internvl3-1B (ours) & 81.11 & 73.06 & 76.87 & \textbf{0.0640} \\
Internvl3-2B (ours) & 81.96 & 73.56 & 77.53 & 0.0659 \\
Internvl3-8B (ours) & \textbf{82.21} & \textbf{76.60} & \textbf{79.31} & 0.2980 \\

\midrule
\multicolumn{5}{c}{\textit{Finetuned with Classifier Head (Ablation)}} \\
\midrule
Internvl3-1B & 61.57 &74.34 &67.36 & 0.0600 \\
Internvl3-2B & 66.48 &75.31	& 70.62 & 0.0612 \\
Internvl3-8B & 68.03 & 71.37 &69.66 & 0.2796 \\

\bottomrule
\end{tabular}
} %
\caption{\scriptsize Performance comparison of baselines and our finetuned models on hallucination detection. The last column refers to the average inference time per sample in seconds. %
Best scores and fastest inference time are in bold.}
\label{tab:main_results}
\vspace{-5mm}
\end{table}

\subsection{Hallucination Detector Module}
\label{sec:results_detection}

\begin{table}[t]
\centering
\resizebox{\textwidth}{!}{%
\begin{tabular}{lcccccccc}
\toprule
\rowcolor{blue!10}
\textbf{Model} & \textbf{ALOHa} $\uparrow$ & \textbf{CHAIRi} $\downarrow$ & \textbf{CHAIRs} $\downarrow$ & \textbf{CLIPScore} $\uparrow$ & \textbf{RefCLIP} $\uparrow$ & \textbf{BERT (Deberta)} $\uparrow$ & \textbf{BERT (D+IDF)} $\uparrow$ & \textbf{Time (s)} $\downarrow$ \\
\midrule

Raw Descriptions & 41.27 & 7.04 & 47.89 & 73.93 & 73.09 & 66.47 & 60.39 & - \\

\midrule
\multicolumn{9}{c}{\textit{Original descriptions + GT hallucinated words + instruction prompt}} \\
\midrule

GPT-5.2 (text-only) & \textbf{43.50} & \cellcolor{green!10}\textbf{0.79} & \cellcolor{green!10}\textbf{5.76} & 73.88 & 73.13 & \textbf{67.32} & \textbf{61.09} & \textbf{0.8152} \\
GPT-5.2 (text + image) & 41.64 & 1.32 & 11.69 & \cellcolor{green!10}\textbf{74.32} & \cellcolor{green!10}\textbf{73.49} & 66.80 & 60.74 & 2.1589 \\

\midrule
\multicolumn{9}{c}{\textit{Original descriptions + predicted hallucinated words + instruction prompt}} \\
\midrule

GPT-5.2 & \textbf{43.27} & \textbf{2.67} & \textbf{19.07} & 73.93 & 73.17 & \textbf{67.29} & \textbf{61.08} & 0.8219 \\
Qwen3-8B & 42.58 & 3.36 & 24.51 & \textbf{74.02} & \textbf{73.23} & 67.11 & 60.94 & 0.3997 \\
Qwen3-14B & 42.69 & 3.33 & 25.08 & 73.97 & 73.16 & 66.67 & 60.55 & 0.7440 \\
Qwen3.5-4B & 42.30 & 3.84 & 28.65 & 73.97 & 73.15 & 66.96 & 60.81 & 0.2697 \\
Qwen3.5-9B & 43.00 & 3.68 & 27.11 & 73.97 & 73.21 & 66.85 & 60.70 & 0.4386 \\
Gemma-4-E4B-it & 42.28 & 2.92 & 22.81 & \textbf{74.02} & 73.21 & 66.54 & 60.45 & 0.3178 \\
Gemma-4-26B-A4B-it & 43.18 & 2.71 & 19.48 & 73.93 & 73.14 & 67.22 & 61.02 & \cellcolor{green!10}\textbf{0.2672} \\

\midrule
\multicolumn{9}{c}{\textit{Image + original descriptions + predicted hallucinated words + instruction prompt}} \\
\midrule

Qwen3.5-4B & 42.15 & 4.50 & 32.71 & 74.02 & 73.31 & 66.68 & 60.60 & 0.5895 \\
Qwen3.5-9B & 42.57 & 3.68 & 27.11 & 74.17 & 73.38 & 66.95 & 60.85 & 0.9129 \\
Qwen3.5-27B & 42.25 & 3.78 & 28.33 & 74.12 & 73.31 & 66.88 & 60.80 & 2.1316 \\
Gemma-4-E4B-it & 42.23 & 3.68 & 28.25 & \textbf{74.22} & 73.37 & 66.72 & 60.63 & \textbf{0.4720} \\
Gemma-4-26B-A4B-it & \cellcolor{green!10}\textbf{43.78} & \textbf{2.77} & \textbf{19.72} & 74.17 & \textbf{73.41} & \cellcolor{green!10}\textbf{67.41} & \cellcolor{green!10}\textbf{61.24} & 0.6154 \\
Gemma-4-31B-it & 42.78 & 3.16 & 23.30 & 74.17 & \textbf{73.41} & 67.13 & 61.00 & 1.9573 \\

\bottomrule
\end{tabular}
}%
\caption{\scriptsize Refinement performance under different input settings. $\uparrow$ indicates higher is better, and $\downarrow$ indicates lower is better. RefCLIP refers to RefCLIPScore and BERT (D+IDF) is Deberta with IDF. Best scores for each setting are in bold. Overall best scores are highlighted in green.}
\label{tab:refinement_results}
\vspace{-7mm}
\end{table}

\cref{tab:main_results} reports the hallucination detection performance 
of all baselines and our finetuned models on the LID test set.

\noindent\textbf{Naive and Zero-Shot Baselines:} The All-None baseline, which predicts \textit{no hallucination} for every sample, achieves an F1 of 43.01, 
establishing a non-trivial lower bound that several zero-shot models struggle to surpass. Among zero-shot models, smaller variants such as InternVL3-1B (41.48) and Qwen2.5-VL-3B (39.77) fail to exceed even this baseline, demonstrating that off-the-shelf MLLMs without task-specific adaptation are insufficient for fine-grained hallucination detection in aerial imagery. Notably, upon manual inspection, InternVL3-2B in the zero-shot setting produces all-None predictions, failing to follow the task instructions entirely, which results in scores identical to the All-None baseline.
Comparatively larger zero-shot models such as InternVL3-8B (49.84) show modest improvement but remain well below acceptable performance. Proprietary GPT models without reasoning achieve low precision despite higher recall, indicating a tendency to generate false positives by flagging non-hallucinated words as hallucinated. Enabling ``low'' reasoning substantially improves performance: GPT-5.4-mini with reasoning reaches an F1 of 70.04, highlighting that reasoning capabilities are beneficial for this task. However, GPT models incur significantly higher inference latency ($>$4 seconds per sample), making them impractical for real-time UAV deployment.

\noindent\textbf{Finetuned Models:} Our supervised finetuned (SFT) models demonstrate 
substantial and consistent improvements over all zero-shot baselines, including the strong GPT reasoning variants. HalLocalizer~\cite{halloc_cvpr25}, finetuned on the same training set, achieves an F1 of 51.71, confirming that discriminative token-level classification is a weaker paradigm for this task compared to our generative prediction approach. Among our SFT models, InternVL3-8B achieves the best overall performance with 82.21 precision, 76.60 recall, and 79.31 F1, outperforming the best zero-shot baseline (GPT-5.4-mini-reasoning) by a margin of over 9 F1 points. Notably, InternVL3-1B achieves the fastest inference at 0.064 seconds per sample while maintaining a competitive F1 of 76.87, making it the most suitable choice for latency-critical deployment. InternVL3-2B offers a strong balance between accuracy (F1: 77.53) and speed (0.066s), providing a practical trade-off for edge deployment scenarios. These results collectively validate that 
finetuning lightweight MSLMs as generative hallucination detectors is both effective and efficient for aerial imagery.

\subsection{Refinement Module}
\label{sec:results_refinement}
\cref{tab:refinement_results} reports the refinement performance across three input settings. Along with the raw or unedited captions, we consider passing three different types of inputs, and have three experimental setups:
 (1) Ground Truth (GT) hallucinated words with and without image, (2) predicted 
hallucinated words without image, and (3) predicted hallucinated words with image. Predicted hallucinated words are from our best performing finetuned detector -- InternVL3-8B.

\noindent\textbf{Reference Settings} (with GT hallucinated words):
GPT-5.2 with text-only input, i.e. without passing the image, achieves the lowest CHAIR scores overall 
(CHAIRi: 0.79, CHAIRs: 5.76), representing reductions of 88.8\% and 88.0\% 
respectively over the original/raw captions (CHAIRi: 7.04, CHAIRs: 47.89), confirming that when hallucinations are perfectly known, a strong LLM can 
correct them with high precision. 
GPT-5.2 with image input reduces CHAIR scores by 81.2\% and 75.6\% (CHAIRi: 1.32, 
CHAIRs: 11.69) over raw descriptions,
while achieving a higher CLIPScore (74.32 vs. 73.88) and 
RefCLIPScore (73.49 vs. 73.13) than the text-only counterpart. This suggests that the image context improves visual grounding and semantic alignment, and also introduces more 
meaningful edits by replacing hallucinated words with visually grounded alternatives rather than simply removing them.
Together, these 
two GPT-5.2 variants establish strong reference baselines for our refinement 
module, with the \textit{raw descriptions serving as the lower bound.}

\noindent\textbf{Text-only Refinement (with predicted hallucinations):}
Among open-source models operating without image context, 
Gemma-4-26B-A4B-it achieves the best hallucination reduction, with CHAIRi 
dropping from 7.04 to 2.71 (61.5\% reduction) and CHAIRs from 47.89 to 
19.48 (59.3\% reduction), alongside an ALOHa improvement of 4.6\% over 
raw descriptions, with the fastest inference time of 0.267 seconds per 
sample, comparable to GPT-5.2 (CHAIRi: 2.67, CHAIRs: 19.07) despite being 
an open-source model. This is particularly noteworthy given that 
Gemma-4-26B-A4B activates only 4B parameters during inference, making it 
an efficient and practical alternative to proprietary models. Qwen3.5-9B 
also performs competitively, reducing CHAIRi by 47.7\% and CHAIRs by 
43.4\% at a low inference cost of 0.439 seconds. The smaller Qwen3.5-4B 
and Gemma-4-E4B-it are the fastest models in this setting (0.270s and 
0.318s respectively), offering a reasonable trade-off between correction 
quality and deployment efficiency and throughput.

\noindent\textbf{Multimodal Refinement (image + predicted hallucinations):}
Incorporating the image as additional context during refinement yields 
further improvements for larger models. Gemma-4-26B-A4B-it achieves the 
best overall performance in this setting, \textbf{reducing CHAIRi by 60.7\% 
(2.77) and CHAIRs by 58.8\% (19.72) over raw descriptions, while achieving 
the highest ALOHa (43.78, a 6.1\% improvement), RefCLIPScore (73.41), 
BERTScore-Deberta (67.41), and BERTScore-Deberta+IDF (61.24)}, demonstrating that 
visual grounding during correction leads to more faithful and semantically 
consistent descriptions. Gemma-4-31B-it also performs competitively, 
reducing CHAIRi by 55.1\% (3.16), but at a significantly higher inference 
cost (1.957s), making Gemma-4-26B-A4B-it the preferred choice for both 
performance and efficiency. Among Qwen3.5 variants, Qwen3.5-9B achieves a 
reasonable balance with a 47.7\% CHAIRi reduction at 0.913 seconds. 
Notably, Gemma-4-E4B-it achieves the \textbf{highest CLIPScore (74.22) and the 
fastest inference (0.472s)} in this setting, making it \textit{the most suitable 
choice for resource-constrained deployment.}

\noindent\textbf{Overall Observations:} Across all settings, all refinement 
models consistently improve over the raw descriptions, 
demonstrating the effectiveness of our 
correction pipeline; these gains stem from conditioning on accurately detected hallucinations, which would not be possible without a reliable detector upstream. 
The text-only and multimodal settings yield comparable 
CHAIR improvements, with the multimodal setting offering additional gains 
in image-similarity metrics (CLIPScore, RefCLIPScore, BERTScore), suggesting 
that visual context helps produce descriptions that are more semantically 
grounded even when hallucination reduction is similar in magnitude.

\subsection{Human Evaluation}
\label{sec:human-eval}

We conduct a human evaluation on 100 randomly selected LID test samples to assess the \textit{faithfulness} and \textit{usefulness} of the refined captions, by comparing the original captions against Gemma-4-E4B-it's refined captions (chosen as the most deployment-suitable model in our experiments). For hallucination assessment, annotators found the refined captions less hallucinated in 80\% of cases, more hallucinated in 1\%, and comparable in 19\%. For usefulness, 73\% of corrections were judged improved, 4\% degraded, 23\% unchanged. This confirms that the refinement step yields genuinely faithful and useful corrections, with detector errors rarely propagating to the refinement stage. Inter-annotator agreement (Cohen's Kappa are 0.84 and 0.83, respectively) indicates almost perfect agreement. The instructions given to the evaluators are in the supplementary (\cref{supple_sec:human_eval}).

\subsection{Ablation Study: Generative vs. Classification Formulation}
\label{sec:ablation}

To validate our design choice for the hallucination detector, we compare our generative formulation against a discriminative baseline, using identical model backbones on the LID dataset. We finetune the same InternVL3 models (\cref{tab:main_results}), replacing the language modelling head with a classification head, under comparable training configurations. As shown in the bottom section of \cref{tab:main_results}, the classification variants achieve markedly lower F1 scores 
compared to our generative approach, 
empirically confirming the advantage of generative prediction.

\subsection{Qualitative Analysis}
\label{sec:qualitative}

\begin{figure}[t]
    \centering
    \includegraphics[width=\linewidth]{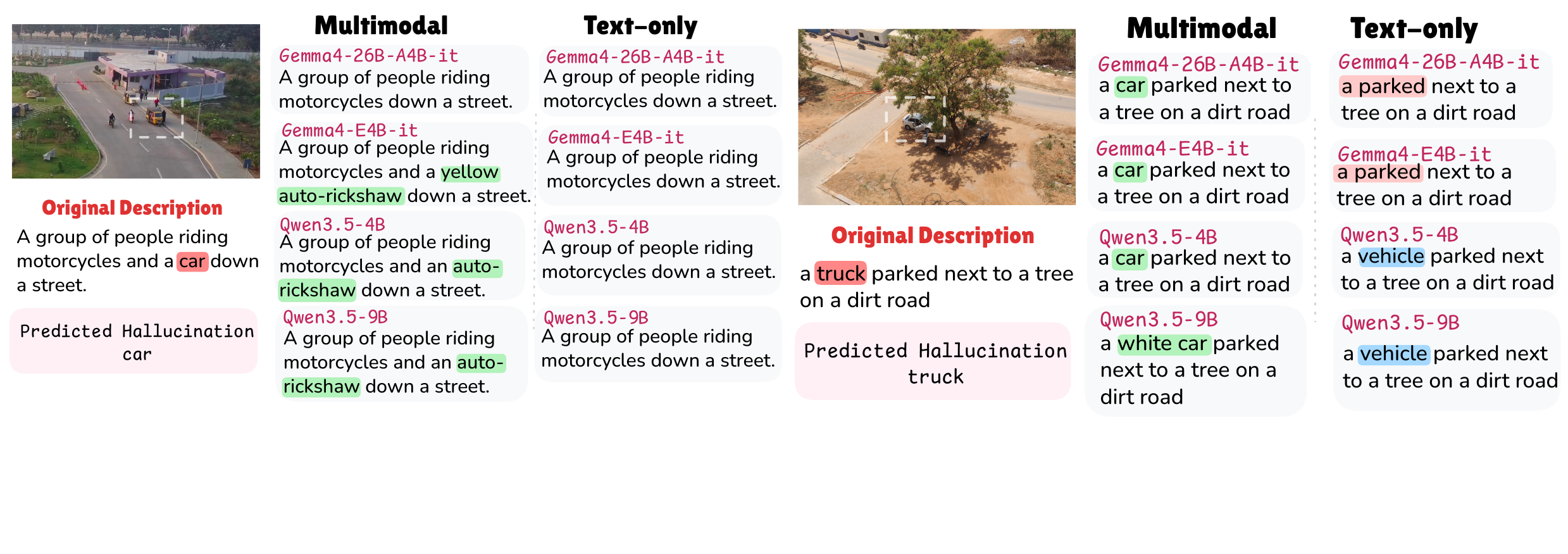}
    \vspace{-15mm}
    \caption{\scriptsize Qualitative Examples: \textit{Multimodal} and \textit{Text-only} refers to the refined captions produced by the Multimodal and Text-only Refinement Modules respectively. The predicted hallucinations are by our Hallucination Detector. Green and red highlights represent the \textit{corrected replacements} and \textit{hallucinated or incorrect words/phrases} respectively. Blue denotes generic or vague substitutions.}
\label{fig:qual_examples}
\vspace{-10pt}
\end{figure}
We present qualitative examples using Gemma-4-26B-A4B-it, Gemma-4-E4B-it, 
Qwen3.5-4B, and Qwen3.5-9B, selected as representative models spanning different sizes 
and architectures from our refinement evaluation.

In ~\cref{fig:qual_examples} (left), the hallucinated word \textit{car} is correctly 
detected, with the actual object being an \textit{auto-rickshaw.} In the multimodal 
setting, models produce visually grounded replacements (\textit{yellow 
auto-rickshaw}, \textit{auto-rickshaw}), while text-only models simply 
delete \textit{car}, yielding complete but less descriptive corrections due 
to the absence of visual context.

\cref{fig:qual_examples} (right) presents a harder case where \textit{truck} is 
hallucinated and the correct object is a \textit{car.} Multimodal models accurately 
substitute it with \textit{car} or \textit{white car}. Text-only models, 
lacking visual context, either produce grammatically incomplete outputs 
(\textit{``a parked next to a tree''}) or resort to vague hypernyms like 
\textit{vehicle}. Together, these examples demonstrate that multimodal refinement enables precise grounded substitutions, whereas text-only correction tends to delete or generalize hallucinated content, potentially reducing description quality.

\subsection{Error Analysis}
\label{sec:error_analysis}

\begin{figure}[b]
\vspace{-4mm}
\centering
\begin{subfigure}[t]{0.45\textwidth}
    \centering
    \includegraphics[width=\linewidth]{{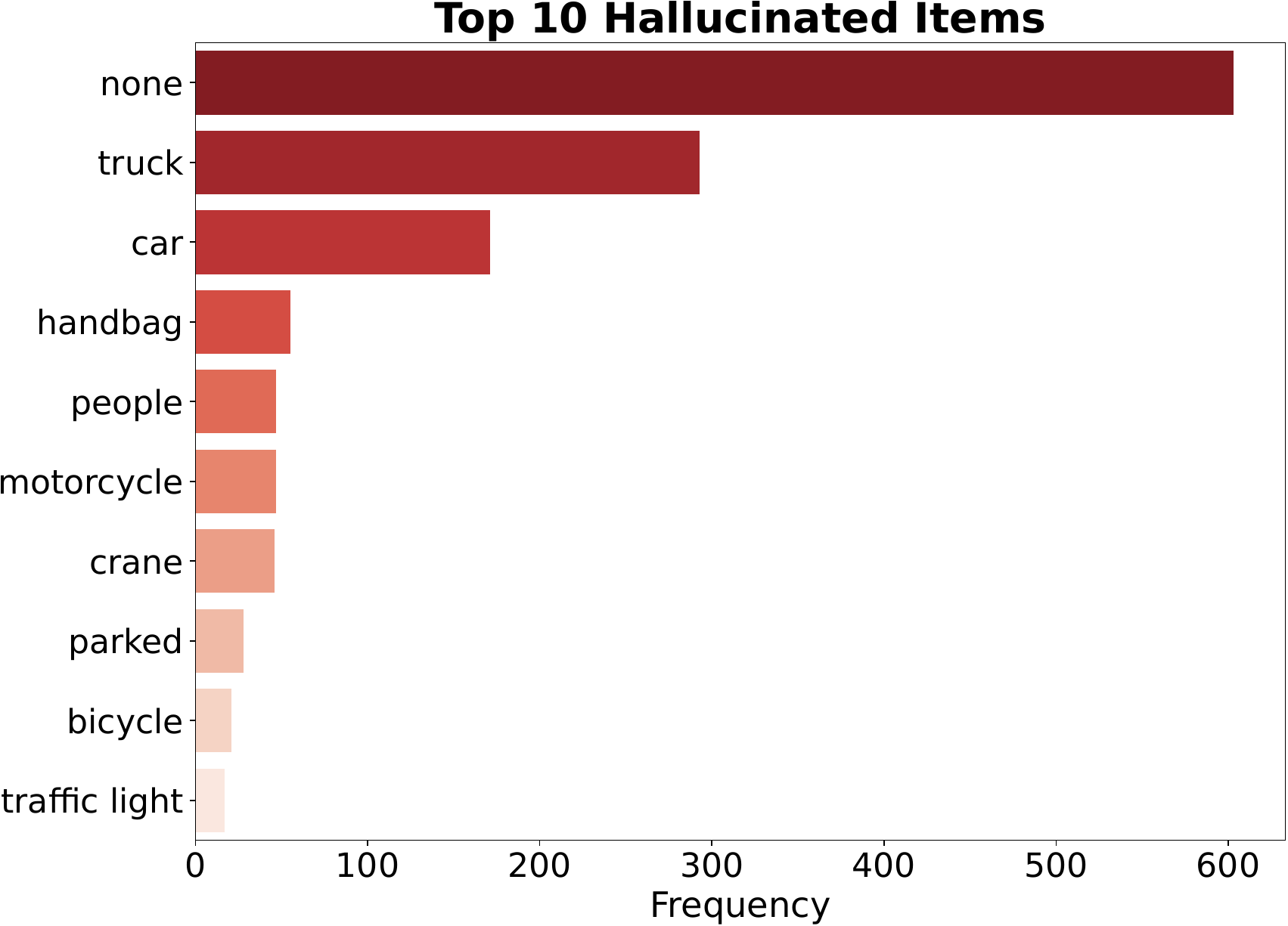}}
    \caption{LID test set}
\end{subfigure}
\hspace{0.04\textwidth}
\begin{subfigure}[t]{0.45\textwidth}
    \centering
    \includegraphics[width=\linewidth]{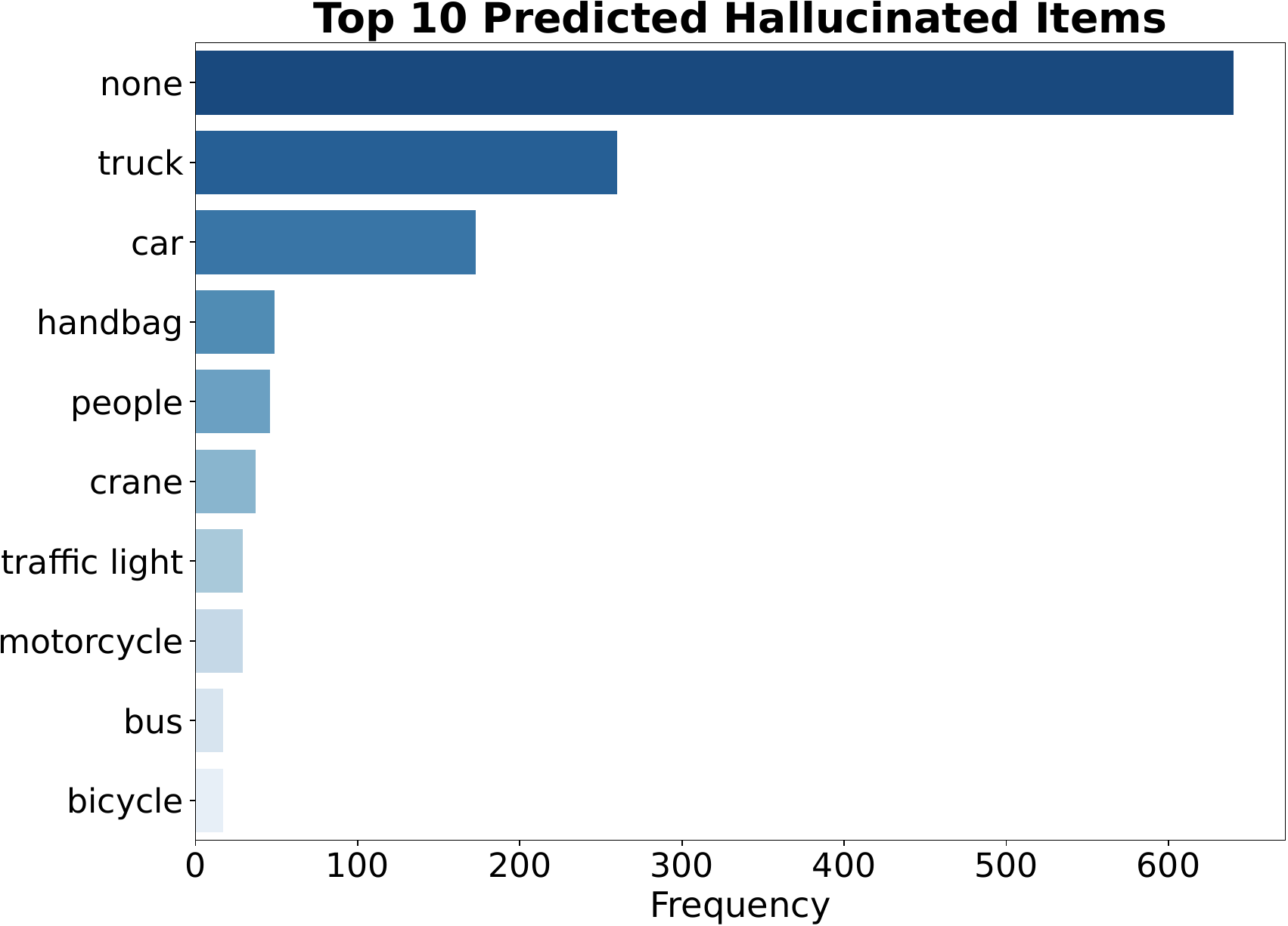}
    \caption{Predicted hallucinations by InternVL3-8B(SFT)}
\end{subfigure}
\vspace{-3mm}
  \caption{\scriptsize Top 10 hallucinated classes in Test vs. Predicted set (all could not be provided due to space constraints). \textit{None} denotes \textit{No hallucination} samples.}
  \label{fig:top15_hal}
  \vspace{-10pt}
\end{figure}

To better understand model behavior, we analyze our best-performing detector, InternVL3-8B (fine-tuned), on the LID test set (1,232 image-caption pairs). It achieves 82.21\% precision, 76.60\% recall, and 79.31\% F1, indicating high precision but some missed hallucinations. Overall, it produces 65.71\% true positives, 14.22\% false positives, and 20.08\% false negatives. At the sample level, it correctly predicts the absence of hallucinations in 90.7\% of the 605 non-hallucinated image-caption pairs, demonstrating strong specificity.

\noindent\textbf{Part-of-Speech (POS) Distribution:} We show a PoS distribution chart in the supplementary (\cref{supple_sec:pos_distribution}). Hallucinated labels in the LID train and test sets span diverse POS, with nouns being most common, followed by verbs, adjectives, adverbs, and pronouns. This highlights that hallucinations extend beyond object-level errors.

\noindent\textbf{Failure Cases:} Errors are common for visually complex objects such as fountains, dump trucks, and fire trucks due to high intra-class variability and visual similarity with related categories. This makes fine-grained distinction in aerial imagery particularly challenging, leading to missed or incorrect detections.

\noindent\textbf{Multi-label Performance:} For 269 multi-label samples, the model predicts all hallucinated labels in 40\% of cases and over half in another 11\%, highlighting the challenge of detecting multiple hallucinations. In contrast, it correctly predicts 84\% of single-label cases, indicating higher difficulty with more labels.

\noindent\textbf{Prediction Patterns:} As shown in \cref{fig:top15_hal}, the most frequently predicted hallucinated categories (e.g., trucks, cars, handbags, people, cranes, traffic lights, motorcycles, buses, and bicycles) is consistent with the ground-truth distribution, indicating that the model captures the dataset's hallucination patterns.

\section{Discussion on Applicability and Deployment}
\label{sec:discussion}
We now discuss the practical feasibility of deploying our framework on real-world UAV platforms, considering both memory and latency.

\noindent\textbf{Memory:} Modern embedded AI systems comfortably support models of the scale used in our framework. For example, the NVIDIA Jetson AGX Orin and Thor provide 64 GB and 128 GB of memory respectively, along with expandable storage 1 TB+, and are widely integrated into autonomous UAV and robotic platforms. Under standard bf16/fp16 inference, our detector models require 
$\approx$ 5 GB, 7 GB, and 19 GB for the InternVL3 1B, 2B, and 8B 
respectively, while the smaller refinement models (Gemma-4-E4B-it, Qwen3.5-4B) operate under 12 GB VRAM. The complete detection and refinement pipeline therefore fits within the memory budget of such platforms, and can be further optimized 
through quantization and TensorRT acceleration. 
While we report larger refinement models as references to benchmark performance, our smaller models achieve comparable results, and our framework allows any model to be substituted based on the available deployment hardware.

\noindent\textbf{Latency:} 
Since our framework operates at the semantic reasoning 
layer rather than time-critical flight control, sub-second to low-second latency 
is sufficient for typical missions such as surveillance and disaster assessment, 
where UAVs capture imagery periodically and generate descriptions for operators. Our fine-tuned detectors operate well within this range, with the InternVL3 1B, 2B, and 8B variants running at 0.064s, 0.066s, and 0.298s per sample respectively. 
Combined with efficient refinement models such as Gemma-4-E4B-it (0.472s), the end-to-end pipeline runs comfortably under 1 second per image.

\section{Conclusion}
\label{sec:conclusion}
In this work, we presented a reference-free framework for fine-grained 
hallucination detection and correction in aerial image captioning. We framed 
hallucination detection as a generative prediction task, to directly predict 
hallucinated words from image-caption pairs in a generator-agnostic manner. 
The detected hallucinations are subsequently corrected by a 
refinement module that conditions on the image, original description, and 
predicted hallucinated words, achieving significant reductions in hallucinations (CHAIR and ALOHa) over the original uncorrected descriptions. Comprehensive experiments across diverse model families and input settings demonstrate the effectiveness of our approach,  while maintaining low inference latency suitable for real-time UAV deployment. 
As future work, we plan to 
extend the framework to video-based UAV captioning 
where temporal consistency introduces additional hallucination challenges.

\bibliographystyle{splncs04}
\bibliography{main}

\clearpage
\appendix
\renewcommand{\theHsection}{supp.\Alph{section}}
\renewcommand{\theHsubsection}{supp.\Alph{section}.\arabic{subsection}}

\renewcommand{\thetable}{S\arabic{table}}
\renewcommand{\theHtable}{supp.\arabic{table}}
\renewcommand{\thefigure}{S\arabic{figure}}
\renewcommand{\theHfigure}{supp.\arabic{figure}}

\title{Supplementary for\\Fine-Grained Generative Hallucination Detection and Correction for Aerial Image Captioning} 

\titlerunning{Fine-Grained Generative Hallucination Detection and Correction}

\author{Debolena Basak\inst{1}\orcidlink{0009-0001-6579-8397} \and
Somsubhra De\inst{2}\thanks{Work done during internship at IIT Hyderabad.}\orcidlink{0000-0003-3839-2575} \and
Maunendra Sankar Desarkar\inst{1}\orcidlink{0000-0003-1963-7338}
}
\authorrunning{Debolena B. et al.}

\institute{$^{1}$IIT Hyderabad, India \quad $^{2}$IIT Madras, India \\
\email{ai20resch11003@iith.ac.in, somsubhrade.04@gmail.com, maunendra@cse.iith.ac.in}}

\maketitle

\vspace{4mm}
\noindent\textbf{Contents}
\begin{itemize}[leftmargin=*, topsep=0.3em, itemsep=0.1em, label={}]
    \item \ref{supple_sec:implementation_details}\quad \nameref{supple_sec:implementation_details}\dotfill \pageref{supple_sec:implementation_details}
    \item \ref{supple_sec:results_hal_only}\quad \nameref{supple_sec:results_hal_only}\dotfill \pageref{supple_sec:results_hal_only}
    \item \ref{supple_sec:prompts_hal_detect}\quad \nameref{supple_sec:prompts_hal_detect}\dotfill \pageref{supple_sec:prompts_hal_detect}
    \item \ref{supple_sec:prompts_refinement}\quad \nameref{supple_sec:prompts_refinement}\dotfill \pageref{supple_sec:prompts_refinement}
    \item \ref{supple_sec:human_eval}\quad \nameref{supple_sec:human_eval}\dotfill \pageref{supple_sec:human_eval}
    \item \ref{supple_sec:confidence_analysis}\quad \nameref{supple_sec:confidence_analysis}\dotfill \pageref{supple_sec:confidence_analysis}
    \item \ref{supple_sec:results_halloc}\quad \nameref{supple_sec:results_halloc}\dotfill \pageref{supple_sec:results_halloc}
    \item \ref{supple_sec:model_choice_details}\quad \nameref{supple_sec:model_choice_details}\dotfill \pageref{supple_sec:model_choice_details}
    \item \ref{supple_sec:pos_distribution}\quad \nameref{supple_sec:pos_distribution}\dotfill \pageref{supple_sec:pos_distribution}
\end{itemize}

\section{Implementation Details}
\label{supple_sec:implementation_details}

We finetune the InternVL3 and Qwen2.5-VL model variants on the LID training set using the MS-Swift framework~\cite{swift}, with LoRA~\cite{hu2021loralowrankadaptationlarge} with rank 16 applied to all linear modules. We use an effective batch size of 8, a learning rate of $1 \times 10^{-5}$ with a warmup ratio of 0.05, and train for 4-5 epochs depending on the model. Training is conducted on two Nvidia RTX 6000 Ada Generation GPUs (48 GB VRAM each). Test-set inferences are performed using one GPU and with a temperature of 0.
Sampling parameters of the refinement models in Tab.~\ref{tab:refinement_results} of the main paper are set to the defaults recommended by each model's official documentation as best practice.\\

\noindent \textbf{HalLocalizer Adaptation and Implementation details:}\\
We use HalLocalizer's \texttt{halloc\_text} configuration, finetuned on the LID training set. It is built on VisualBERT and takes both the image and the caption as input. We do not use the other variants, as they require access to the caption generator's internal states, which is incompatible with our generator-agnostic setting.\\
\textbf{Label conversion:} for each LID caption, we map the annotated hallucinated words/labels into its corresponding character-level spans (\texttt{start:end}) following HalLocalizer's annotation format. This pre-processing ensured that the LID data were compatible with Hallocalizer's pipeline.\\
\textbf{Evaluation:} For each token, HalLocalizer outputs scores for the hallucinated and non-hallucinated classes, and we take the class with the higher score (argmax over the output logits) as the prediction, so no threshold tuning is required. The token-level predictions are then converted to word level: subword tokens are merged into whole words, and a word is marked as hallucinated if any of its subwords is predicted as hallucinated. Precision, Recall and F1 are then computed against the LID ground truth as described in Sec.~\ref{sec:eval} of the main paper.

\section{Detector Performance on Hallucination-only samples}
\label{supple_sec:results_hal_only}

To decouple localization performance from the easier ``None'' decision, we evaluate the hallucination detectors' performances on the subset of LID test samples that contain actual hallucinations. 
We recompute Precision, Recall and F1 scores for the hallucination-only samples. We use the open-sourced models of Tab.~\ref{tab:main_results} of main paper for this comparison. Results are shown in \cref{tab:hal-only_results}.
Our finetuned models outperform HalLocalizer by up to \textbf{54.26 F1} points and consistently surpass the zero-shot counterparts, showing that the localization advantage of our generative formulation holds even when the influence of the null decision is removed.

\begin{table}[h]
\centering
\resizebox{0.6\columnwidth}{!}{%
\begin{tabular}{lccc}
\toprule
\rowcolor{blue!10}
\textbf{Model} & \textbf{Precision$\uparrow$} & \textbf{Recall$\uparrow$} & \textbf{F1$\uparrow$} \\
\midrule
\multicolumn{4}{c}{\textit{Zero-shot baselines}} \\
\midrule

InternVL3-1B             & 2.33  & 1.54  & 1.85  \\
InternVL3-2B             & 0     & 0     & 0     \\
InternVL3-8B             & 17.70 & 12.30 & 14.51 \\
Qwen2.5VL-3B             & 39.14 & 57.79 & 46.67 \\
\midrule
\multicolumn{4}{c}{\textit{Finetuned Models}} \\
\midrule
HalLocalizer~\cite{halloc_cvpr25} & 21.83 & 14.94 & 17.74 \\
Qwen2.5VL-3B (ours)      & \textbf{78.46} & 66.80 & \textbf{72.16} \\
InternVL3-1B (ours)      & 75.25 & 62.91 & 68.53 \\
InternVL3-2B (ours)      & 73.40 & 61.07 & 66.67 \\
InternVL3-8B (ours)      & 76.71 & \textbf{67.83} & 72.00 \\
\bottomrule
\end{tabular}
}
\vspace{2mm}
\caption{Hallucination detection scores for hallucination-only samples corresponding to Tab.~\ref{tab:main_results} of the main paper.}
\label{tab:hal-only_results}
\end{table}

\clearpage

\section{Prompts used for Hallucination Detector Module}
\label{supple_sec:prompts_hal_detect}
\vspace{2mm}

\begin{tcolorbox}[size=fbox,colback=lightgray!20!white, colframe=black, width=0.9\linewidth, arc=1pt, center, fontupper=\small, fonttitle=\small, title= Prompt for InternVL3 finetuning]

You are InternVL3, a multimodal model that reads images and text. Your task is to detect hallucinated words in image descriptions. Hallucinations are objects, counts, positions, attributes, relationships, or actions/details not present in the image. Ignore subjective interpretations.\\

\textbf{Input:} An image and a candidate description.\\
\textbf{Output format:}\\
- ``None'' if the description fully matches the image.\\
- Otherwise, a comma-separated list of hallucinated words/phrases.\\
Do not add any extra commentary. Be conservative - only flag clear visual inconsistencies.\\
\textbf{User Content:}\\
Image: \texttt{<image>} \\
Description: \texttt{<description>}\\
Find hallucinated words from the description that don't match the image.
\end{tcolorbox}

\vspace{4mm}
\begin{tcolorbox}[size=fbox,colback=lightgray!20!white, colframe=black, width=0.9\linewidth, arc=1pt, center, fontupper=\small, fonttitle=\small, title=Example Prompt for Zeroshot Baselines]
You are InternVL3, a multimodal model that reads images and text, and detects hallucinated words in image descriptions. Hallucinations are objects, counts, positions, attributes, relationships, or actions not present in the image. Ignore subjective interpretations.\\

\textbf{Input:} An image and a candidate description.\\
\textbf{Output format:} \\
- ``None'' if the description fully matches the image.\\
- Otherwise, a comma-separated list of hallucinated words. \\
Do not add extra commentary. Be conservative - only flag clear visual inconsistencies. Return only minimal hallucinated noun phrases from the description. Do not include counts, locations, adjectives, or explanations unless they are themselves the hallucinated item.\\
\textbf{User Content:}\\
Image: \texttt{<image>}\\
Description: \texttt{<description>}\\
Find hallucinated words from the description that don't match the image.
\end{tcolorbox}

\clearpage

\section{Instruction Prompts used for Refinement Modules}
\label{supple_sec:prompts_refinement}
\vspace{2mm}

\begin{tcolorbox}[size=fbox,colback=lightgray!20!white, colframe=black, width=0.9\linewidth, arc=1pt, center, fontupper=\small, fonttitle=\small, title= System Prompt for Multimodal Refinement Module]
    You are a precise description editor. You are given an aerial image, its text description, and a list of suspected hallucinated words predicted by an external model (which may not always be accurate).\\

    Your task is to produce a corrected description by following these rules:
    \begin{enumerate}[leftmargin=*, topsep=0.1pt]
        \item Use the image as the primary source of truth.
        \item Use the suspected hallucinated words as hints — verify each against the image before removing or modifying.
        \item Make only minimal edits to the original description — remove or rephrase only the specific words, phrases, or sentences that are clearly not visible in the image.
        \item Do not rewrite, restructure, or paraphrase the description beyond what is necessary to remove the hallucination.
        \item Do not add any new details, objects, or observations from the image that are not already present in the original description.
        \item Preserve the original wording, style, and structure as much as possible.
        \item Output only the minimally corrected description.\\
    \end{enumerate}
    
    \textbf{User Content:}\\
    Description: \texttt{<Original Description>}\\
    Suspected Hallucinated words: \texttt{<Predicted Hallucinations>}\\
    Correct the description based on the image and the suspected hallucinated words above.\\
\end{tcolorbox}

\vspace{4mm}
\begin{tcolorbox}[size=fbox,colback=lightgray!20!white,colframe=black, width=0.9\linewidth, arc=1pt, center, fontupper=\small, fonttitle=\small, title= System Prompt for Text-only Refinement Module]

 You are given a description of an aerial view image and a set of hallucinated words. Your task is to correct the description by removing the hallucinated words. Modify or delete the sentences or phrases involving the hallucinated words' contexts. Do not add any additional details. Output only the modified description.\\

 \textbf{User Content:} \\
 Description: \texttt{<Original Description>}\\
 Hallucinated words: \texttt{<Predicted Hallucinations>}\\
\end{tcolorbox}

\clearpage

\section{Human Evaluation}
\label{supple_sec:human_eval}
\begin{figure}[h]
\centering
\begin{subfigure}[t]{0.9\textwidth}
    \centering
    \includegraphics[width=\linewidth]{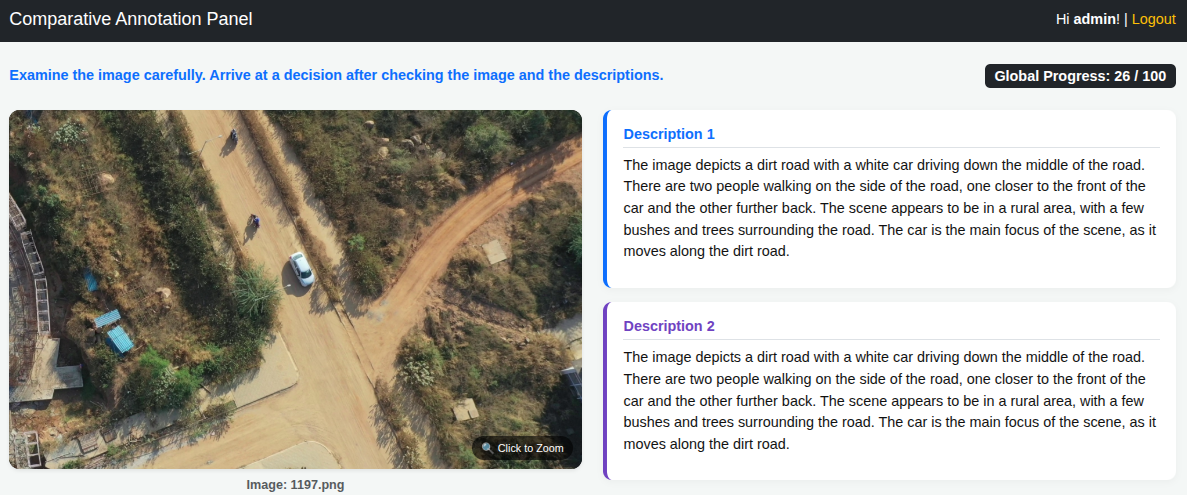}
\end{subfigure}

\begin{subfigure}[t]{0.9\textwidth}
    \centering
    \includegraphics[width=\linewidth]{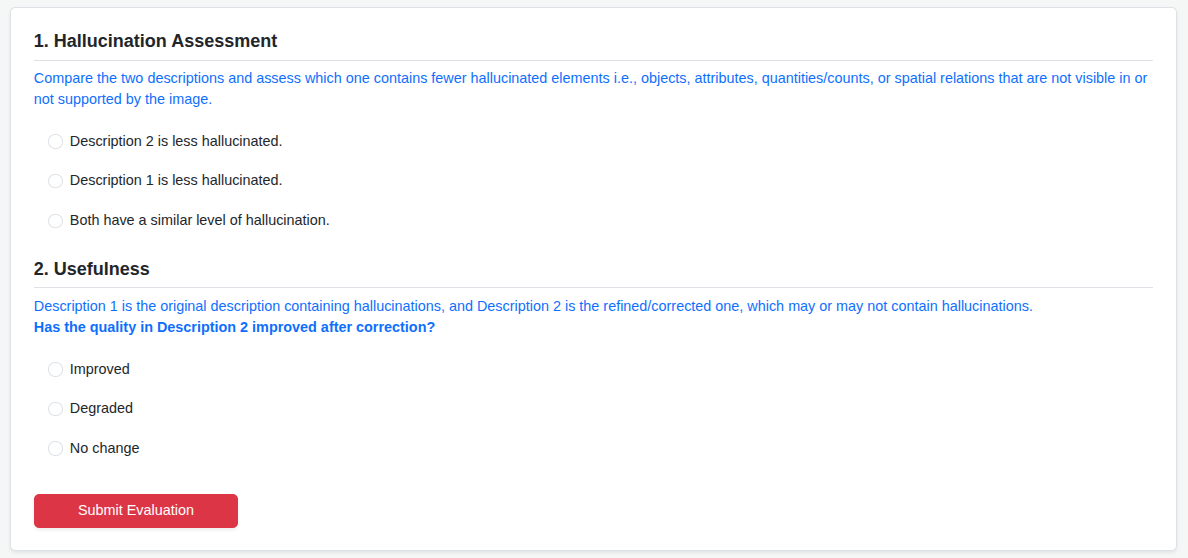}
\end{subfigure}
\caption{Human annotation interface}
\label{fig:annotation_tool}
\end{figure}

Human evaluation on 100 samples was conducted by two undergraduate engineering students. Using an annotation tool, they assessed the faithfulness and usefulness of the refined captions given the original captions and the image (with a zoom option). The instructions and a sample from the annotation tool are shown in \cref{fig:annotation_tool}.

\section{Association of Detector Confidence with Correction Quality}
\label{supple_sec:confidence_analysis}

Our generative detector exposes token probabilities for each predicted hallucinated span as a byproduct. We study whether this confidence is associated with the quality of the resulting corrected descriptions.

\noindent\textbf{Setup:} We define the \emph{span confidence} as the length-normalised geometric mean of the token probabilities of the predicted span. We use the human evaluation samples of Sec.~\ref{sec:human-eval} of the main paper for which the detector predicted a hallucination, since captions with a ``None'' prediction are left unchanged. We rank these samples by span confidence and split them into four quartiles, where Q1 and Q4 are the lowest and highest confidence quartiles. We then compare each quartile with the human judgements of the same samples.

\noindent\textbf{Results:} The rate of corrections judged ``improved'' rises from 52\% in Q1 to 92\% in Q4. The rate of corrections judged ``less hallucinated'' rises from 68\% in Q1 to 84\% in Q4. We also treat each human outcome as a binary label and use span confidence as the ranking score. 
The ROC-AUC is 0.623 for hallucination reduction and 0.733 for usefulness.
Higher-confidence predictions are therefore more likely to receive positive human judgements.
These results provide evidence that detector confidence is associated with correction quality.

\section{Hallucination Detection on Additional Dataset}
\label{supple_sec:results_halloc}

Given the lack of an aerial dataset with fine-grained hallucination annotations, our study primarily focuses on the only such available LID dataset. Nonetheless,
to demonstrate the generalization of our hallucination detection formulation, we use the general-view \texttt{HalLoc-Caption} dataset~\cite{halloc_cvpr25} for further evaluation.
We finetuned the InternVL3 detectors (as they are the best models and core contributions of Tab.~\ref{tab:main_results} of main paper), and \textit{HalLocalizer}~\cite{halloc_cvpr25} on the training set, using the same protocol as for LID, and evaluated them on the test set. 
We use the same \texttt{train}, \texttt{val}, and \texttt{test} splits of \texttt{HalLoc-Caption} data, as provided by the authors~\footnote{\href{https://huggingface.co/datasets/uunicee/HalLoc/tree/main}{HalLoc Dataset link}}. Results are given below in \cref{tab:halloc_detection}. 

\begin{table}[h]
\vspace{-4mm}
\centering
\resizebox{0.6\columnwidth}{!}{%
\setlength{\tabcolsep}{4pt}
\begin{tabular}{lccc}
\hline
\textbf{Model} & \textbf{Precision$\uparrow$} & \textbf{Recall$\uparrow$} & \textbf{F1$\uparrow$}\\
\hline
HalLocalizer~\cite{halloc_cvpr25} & 76.46 & 71.97 & 74.14\\
InternVL3-1B (ours) & 80.51 & 69.69 & 74.71 \\
InternVL3-2B (ours) & \underline{86.53} & \underline{80.35} & \underline{83.33}\\
InternVL3-8B (ours) & \textbf{89.41} & \textbf{83.29} & \textbf{86.24}\\
\hline
\end{tabular}
}
\vspace{2mm}
\caption{Performance of Hallucination Detectors on HalLoc-Caption data split.}
\vspace{-4.5mm}
\label{tab:halloc_detection}
\end{table}

The trend seen in Tab.~\ref{tab:main_results} of main paper is preserved: Our detectors outperform HalLocalizer in F1, and performance scales with model size.  Scores on the HalLoc dataset are higher than the corresponding LID scores for InternVL3-2B, 8B and HalLocalizer, re-affirming challenges in aerial-view hallucination detection. 
The gap to HalLocalizer is also much wider on LID (27.6 F1 points for InternVL3-8B) than on HalLoc (12.1 F1, \cref{tab:halloc_detection}). 
The observations reinforce the contribution of our generative formulation and evaluations, and demonstrate the usefulness of UAV domain-specific detectors.

\section{Model Choice Details}
\label{supple_sec:model_choice_details}

\textbf{Hallucination Detector Models:}
For the hallucination detection task, we fine-tune popular lightweight instruction-tuned Multimodal Small Language Models (MSLMs) that offer strong vision-language understanding while maintaining fast inference speed, making them suitable for real-time UAV deployment.
Specifically, we finetune InternVL3-Instruct (1B, 2B, 8B)~\cite{internvl3} and Qwen2.5-VL-Instruct-7B~\cite{qwen2.5vl} using LoRA~\cite{hu2021loralowrankadaptationlarge}.
We select these model families for their strong multimodal capabilities and distinct architectural and training strategies. 
InternVL3~\cite{internvl3} uses native multimodal pre-training with joint parameter optimization, pixel-unshuffle for scalable high-resolution processing, and Variable Visual Position Encoding (V2PE). It also uses SFT and Mixed Preference Optimization (MPO) for post-training. Qwen2.5-VL~\cite{qwen2.5vl} supports dynamic image resolutions, native aspect ratios, and 2D-RoPE to preserve spatial structure. Qwen2.5-VL further uses absolute pixel coordinates for bounding boxes and trains a native dynamic-resolution ViT from scratch.
The architectural enhancements are intended to make the selected models particularly strong for vision-centric and multimodal understanding tasks.\\
\noindent\textbf{Refinement Module:} As described in~\cref{sec:refinement} of the main paper, we employ advanced pre-trained MLLMs or MSLMs to correct hallucinated content (if detected by our detector module) and produce refined descriptions via an instruction system prompt. We select model families with strong instruction-following and multimodal capabilities: 
Qwen-3.5-(4B, 9B, 27B)~\cite{qwen35blog} and Gemma-4-it (E4B, 26B-A4B, 31B)~\cite{gemma_4_2026}. These are natively multimodal and instruction-tuned variants with built-in reasoning capabilities and configurable reasoning modes.
Qwen-3.5 employs a unified vision-language foundation 
via early fusion of multimodal tokens for better visual and cross-modal understanding. 
It has an efficient hybrid architecture consisting of Gated Delta Networks combined with sparse Mixture-of-Experts (MoE) that delivers high-throughput inference with minimal latency and cost overhead~\cite{qwen35blog}. 
Gemma-4~\cite{gemma_4_2026} offers complementary dense and MoE variants with native support for images at variable resolutions, enabling reliable visual grounding during correction. 
Notably, the smaller variants Qwen3.5-4B and Gemma-4-E4B, along with the Gemma-4-26B-A4B MoE model, which activates only 4B parameters during inference, are specifically designed for efficient local execution, making them suitable for resource-constrained deployment scenarios.
For the text-only refinement setting, we also experimented with GPT-5.2~\cite{gpt52}, Qwen3-8B, and Qwen3-14B. 
Since Qwen3.5 and Gemma-4 variants demonstrate a superior cost-performance 
trade-off, we extend their evaluation to both text-only and multimodal settings 
for a comprehensive comparison across input modalities. Together, these two 
architecturally distinct model families allow us to benchmark the effectiveness 
of our zero-shot refinement approach across diverse multimodal backbones.

\clearpage

\section{Distribution of PoS in the LID dataset}
\label{supple_sec:pos_distribution}

\begin{figure}[h]
\centering
  \includegraphics[width=0.8\textwidth]{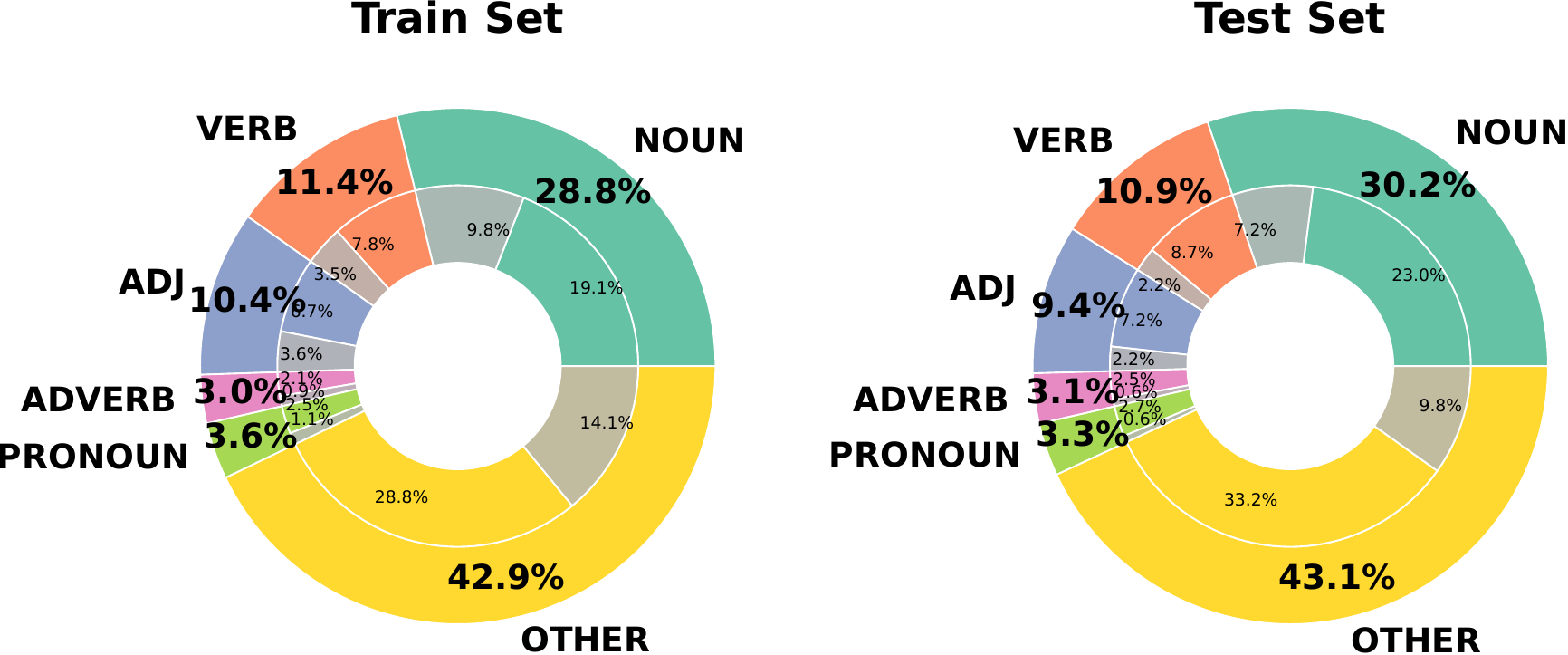}
  \caption{Distribution of PoS in the LID dataset \textit{(The grey portion in the inner ring represents the non-hallucinated patch.)}}
  \label{fig:pos1}
\end{figure}

We provide a donut chart to show the distribution of the different parts-of-speech in the LID dataset in \cref{fig:pos1}.

\end{document}